\documentclass[10pt,twocolumn,letterpaper]{article}

\usepackage{wacv}
\usepackage{times}
\usepackage{epsfig}
\usepackage{graphicx}
\usepackage{amsmath}
\usepackage{amssymb}
\usepackage{booktabs}
\usepackage{float}

\usepackage{tabularx}
\usepackage{multirow}
\usepackage{subcaption}
\usepackage{rotating}

\usepackage{amssymb}
\usepackage{pifont}
\newcommand{\cmark}{\ding{51}}%
\newcommand{\xmark}{\ding{55}}%
\usepackage{enumitem}

\usepackage[accsupp]{axessibility}
\usepackage[font=small,skip=2pt]{caption}

\newcommand\numberthis{\addtocounter{equation}{1}\tag{\theequation}}

\def\wacvPaperID{} 

\wacvapplicationstrack 

\wacvfinalcopy 

\ifwacvfinal
\usepackage[breaklinks=true,bookmarks=false]{hyperref}
\else
\usepackage[pagebackref=true,breaklinks=true,colorlinks,bookmarks=false]{hyperref}
\fi

\begin{document}

\title{Hand Guided High Resolution Feature Enhancement for Fine-Grained Atomic Action Segmentation within Complex Human Assemblies}

\author{Matthew Kent Myers\\
\small School of Engineering\\
\small Newcastle University\\
{\tt\small M.J.Kent-Myers2@ncl.ac.uk}
\and
Nick Wright\\
\small School of Engineering\\
\small Newcastle University\\
\and
A. Stephen McGough\\
\small School of Computing\\
\small Newcastle University\\
\and
Nicholas Martin\\
\small Tharsus Ltd.\\
}

\maketitle
\thispagestyle{plain}
\pagestyle{plain}

\begin{abstract}
Due to the rapid temporal and fine-grained nature of complex human assembly atomic actions, traditional action segmentation approaches requiring the spatial (and often temporal) down sampling of video frames often loose vital fine-grained spatial and temporal information required for accurate classification within the manufacturing domain. In order to fully utilise higher resolution video data (often collected within the manufacturing domain) and facilitate real time accurate action segmentation  -- required for human robot collaboration -- we present a novel hand location guided high resolution feature enhanced model. We also propose a simple yet effective method of deploying offline trained action recognition models for real time action segmentation on temporally short fine-grained actions, through the use of surround sampling while training and temporally aware label cleaning at inference. We evaluate our model on a novel action segmentation dataset containing 24 (+background) atomic actions from video data of a real world robotics assembly production line. Showing both high resolution hand features as well as traditional frame wide features improve fine-grained atomic action classification, and that though temporally aware label clearing our model is capable of surpassing similar encoder/decoder methods, while allowing for real time classification.

\end{abstract}

\section{Introduction}

\begin{figure}
    \hspace*{-.75cm} 
    \centering
    \includegraphics[width=.53\textwidth]{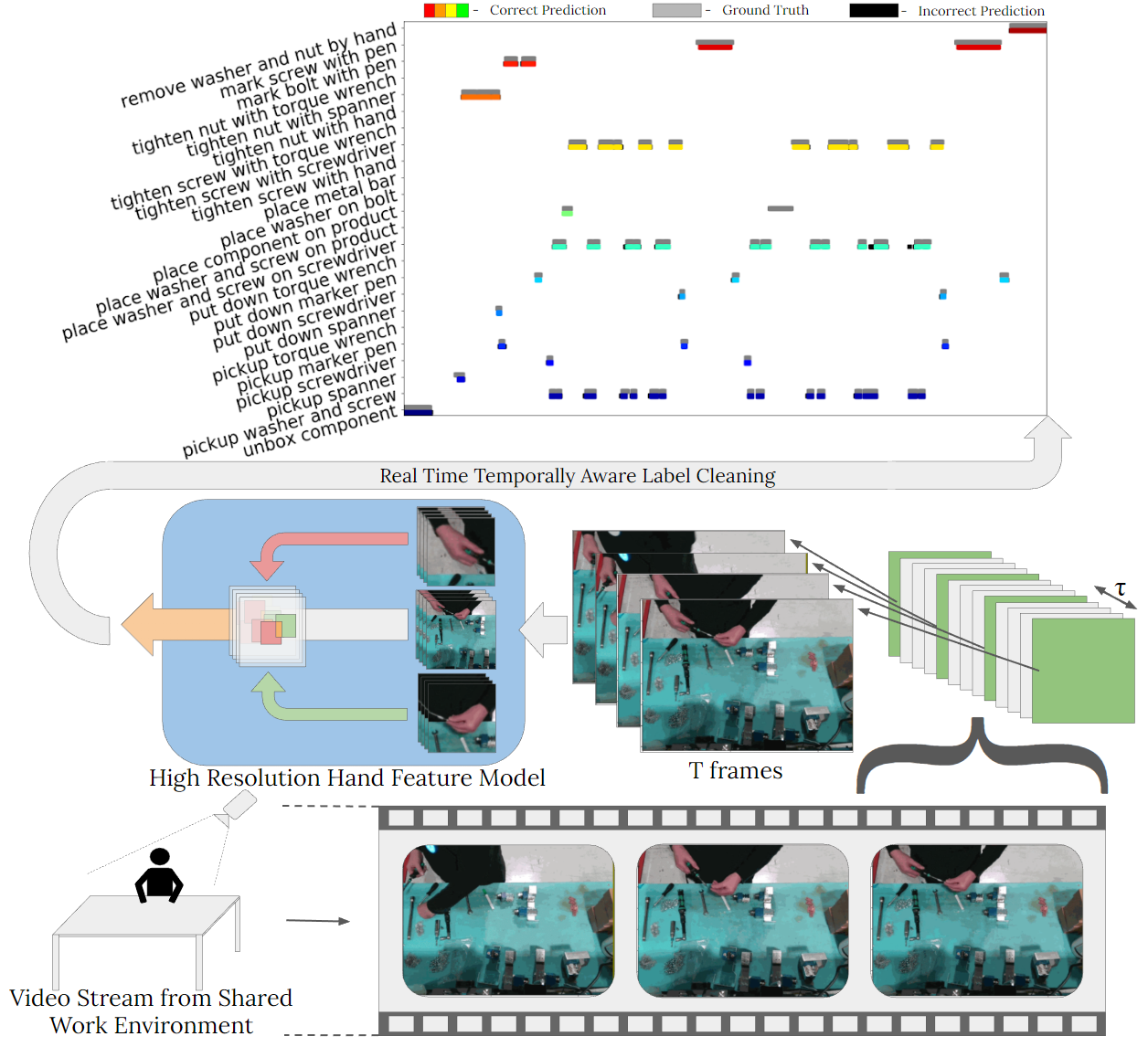}
    \caption{\small Overview of our action segmentation approach to classifying real world assembly data. We operate temporally aware label cleaning in an overlapping sliding window fashion, utilising a temporally aware high resolution hand feature enhanced frame wide backbone model.}
    \label{fig:intro}
\end{figure}

While advances in robotics have made significant progress in automating many simple and repetitive labour heavy tasks within manufacturing, many complex manufacturing assemblies requiring assembly adaptability or robustness still require human operators. Human-Robot Collaboration (HRC) has been identified as a key component in moving towards a smart assembly procedure in the next few years~\cite{cyber-physical_production}. Where humans and robots can work in tandem on assemblies, leading to increased productivity, reliability, more attractive working conditions \cite{symbiotic_human-robot}, as well create a plethora of quality assurance and guidance tools (\eg \cite{InteractiveWA,DuploTrack}) reducing human error and speeding up assembly. One of the core challenges of implementing successful HRC systems is the ability for machines to understand the environment around them in a quick and robust manner; with a successful HRC system requiring the ability to understand an assembler's actions and intentions in real time in order to intelligently assist them and provide an efficient hybrid work system~\cite{morphology}.

Motivated by recent atomic-action datasets (\eg \cite{haa500,Epic-Kitchens,Breakfast,charades}) and the significant advances in fine-grained action recognition (AR) the utilisation of video data captured within the shared work environment is now an attractive, unintrusive, and information rich approach to classifying fundamental atomic actions between a worker, product, part or tool. With the accurate classification of these actions allowing assemblies to be understood as a nested set of fundamental atomic actions. Whilst many works have focused on classifying actions within manufacturing environments, most of these works have been limited in their approach; often operating within a controlled environment, classifying coarse actions with single frames, or performing action segmentation in an offline manner. In contrast, many HRC systems require real time classification of spatially and temporally similar fine-grained actions, which is often very challenging due to the spatial limitations of AR inputs (224x224) where often significant downsampling is used -- removing vital information collected by the camera.

With this motivation in mind we develop a novel AR model facilitating the use of enhanced spatial features via hand guided high resolution feature enhancement, capable of operating in real time in a sliding window fashion with temporally aware label cleaning for action segmentation, as shown in Fig. \ref{fig:intro}. Our novel model architecture is evaluated on a domain specific novel assembly dataset collected from a real world robotics sub assembly manufacturing production line. We perform extensive experimentation on extended sequences of unseen video data and provide a class specific performance analysis, revealing the importance of enhanced hand features for improving fine-grained atomic action classification. Our main contributions are as follows:

\begin{itemize}[noitemsep]
    \item We Introduce a novel AR model utilising hand location guided high resolution feature extraction to allow fine grained details to aid a backbone AR model and show both high resolution and frame wide features are key to accurate classification.
    \item We show that running a temporally aware model trained on short video clips from our domain dataset can be successfully applied in a simple sliding window via the use of surround sampling during training and temporally aware label cleaning, surpassing the performance of similar encoder/decoder methods, whilst operating in real time.
    \item We show that the performance of encoder/decoder methods can be improved significantly, and training time reduced, via the use of in domain feature extraction via model training under our framework. 
\end{itemize}

\section{Related Work}

\subsection{Action Segmentation}
Current action segmentation approaches can be broadly broken down into two stages: An initial feature extraction process (often on a frame or short segment level) followed by a frame-wise sequence prediction stage, often facilitating the ability to learn longer term temporal dependencies to refine action boundaries and reduce over segmentation.

\subsubsection{Feature Extraction}
Feature extraction generally falls into two main categories: 2D or 3D Convolutional Neural Network (CNN) based methods. 2D CNN methods utilise a 2D CNN \cite{resnet,AlexNet,Inception} backbone to extract frame level predictions with a temporal aggregation step, such as a simple averaging \cite{TSN} or a more complex operation such as an LSTM block \cite{LSTM_vid_class} to produce features for a set of input frames. In contrast, 3D CNN based methods (\eg \cite{3D_resnet,P3D,factorised,C3D,r(2+1)d}) expand the 2D convolutional operation into the temporal dimension, performing 3D convolutions across space and time, extracting joint spatial and temporal semantics from voxels of raw video data. Other works utilise the inherent computational efficiency of 2D CNNs via the incorporation of temporally aware modules inserted within 2D CNN architectures (\eg \cite{TSM,GST}) facilitating the mixing of features across frames, allowing for a computationally efficient and parameter light methods of facilitating temporal learning. 

Our work is similar in motivation to many works focusing on improving feature extraction via lateral connections between different layers of a model \cite{feature_pyramid, temporal_pyramid}, different models operating on different frame rates \cite{slowfast} or modalities \cite{fuse}. In contrast to these approaches which all utilise a single fixed input resolution, we utilise a secondary stream with access to higher resolution image information which is used to enhance the backbone frame wide features.

\vspace{-3mm}
\subsubsection{Sequence Predictions}
In order to produce frame level predictions across a long untrimmed video many temporal modelling methods have been used. Most simply, a sliding window \cite{cooking}, often operating with a recurrent network \cite{recurrent} or a feature buffer \cite{movienets,TSM} to allow for arbitrary length temporal modelling in an online manner, can be utilised to make sequence predictions. Many approaches replace the relatively short term temporal modelling capabilities of recurrent models with temporal convolutions \cite{conv_vs_reccurrent}, improving action boundaries and minimising over segmentation errors via encoder/decoder methods \cite{ms-TCN,TCN} typically operating on precomputed features extracted by a pretrained 3D CNN model. In contrast, we show show through the use of surround sampling while training on in domain data, a simple temporally aware label cleaning method at inference can outperform more computationally expensive encoder/decoder methods operating on kinetics \cite{kinetics} pretrained features, while performing real time action segmentation.

\subsection{Assembly Understanding}

\begin{figure*}[h]
\begin{center}
\includegraphics[width=\textwidth]{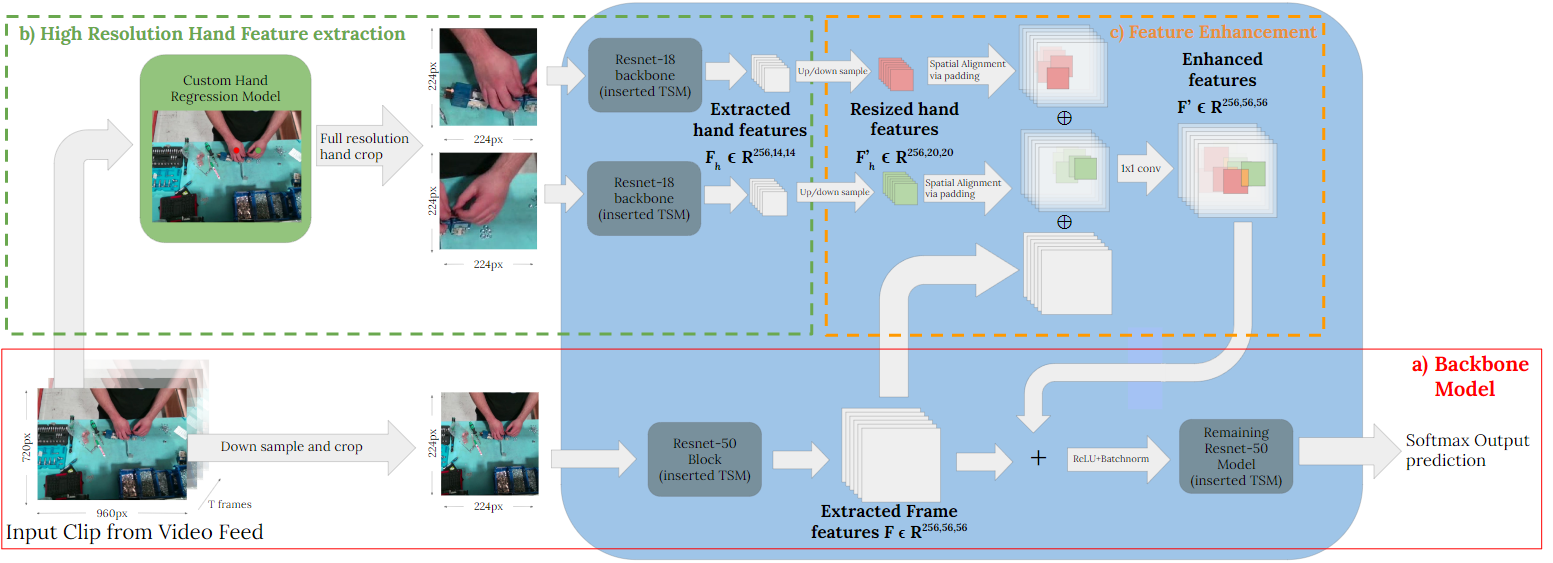}
\end{center}
   \caption{An overview of our proposed method of utilising high resolution hand features to improve frame level features. Our method can be broken into three main sections: a) A backbone model extracting frame wide features from a set of spatially downsampled video frames. b) Feature extraction from higher resolution images focused on locations extracted via a hand regression model. c) Backbone model feature enhancement via the spatially aware combination of extracted features.}
\label{fig:overview}
\end{figure*}

Since the breakthrough of the CNN during the 2010s, there has been fairly limited work applying AR models to the manufacturing domain in order to understand and facilitate HRC. Wang \etal \cite{WANG201817} fine-tune two AlexNet\cite{AlexNet} models to classify three parts being used as well as three current ``motions" (grasping, holding and assembling) during an engine block assembly within a controlled environment. Al-Amin \textit{et al.} \cite{multi-sensor} utilise a multi-modal sensor fusion, including image data, to classify seven actions within a controlled assembly process. Zhang \textit{et al.} \cite{bi-cnn} utilise a bi-stream CNN to extract human and object information from video frames, with a variable-length Markov model making future action predictions within a controlled assembly scenario. Xiong \etal \cite{XIONG2020605} incorporates motion information via the use of a modest two stream network classifying actions occurring within an engine block assembly video scraped from YouTube. Jones \etal \cite{Lego} develop a framework for understanding fine-grained assembly actions as a structure. In contrast to these works, which are all limited in approach, and are deployed within controlled environments, our work utilises fully spatio-temporal feature extraction and is applied to our novel real world assembly dataset.

As human assemblies naturally utilise hands, there are several works with similar motivations to ours. Lui \etal \cite{hand_motion} utilise a modest 3D CNN operating on fixed length segments composed of cropped hand images to classify real world assembly data into seven classes. Kobayashi \etal \cite{hand_attention} implement a similar method however note that such inputs remove much of the contextual surrounding information which is often required for accurate classification. As such in the same work Kobayashi \etal introduce an attention based model where pose features from single video frames are used to draw attention to hands on real world assembly data. Images are classified into eleven actions (such as take product, put on component, grab driver) using an encoder/decoder for action segmentation. In contrast, we show that using higher resolution pixel data from around the hands to enhance full frame features provides more information than drawing attention to hands during assemblies, whilst allowing for full frame features to provide context when classifying, unlike hand cropping methods.

\section{Proposed Method}
The following section introduces our proposed method for action segmentation within the manufacturing domain. Our method aims to specifically tackle the challenge of distinguishing between spatially fine-grained actions, where two temporally similar actions \eg $\textit{tighten screw with torque wrench}$ $\textit{vs.}$ $\textit{tighten nut with torque wrench}$ require fine-grained spatial detail to separate. As the vast majority of AR approaches are based on image classification CNNs requiring a fixed size input of 224x224 pixels, high resolution video frames (typically upwards of 720p) must be downsampled before being input into the model, removing vital high resolution information required to distinguish between small or similar looking components/parts. In order to address this issue our method outlined in Fig. \ref{fig:overview} utilises high resolution data extracted around the assembler's hands (where most relevant fine grained information: tool, component, fixings \etc is typically concentrated) to enhance the features extracted by a backbone frame wide model. Our model can be separated into 3 main section: a) a backbone frame wide feature extractor, b) high resolution hand feature extraction and c) back bone feature enhancement via a spatially aware feature enhancement module, discussed in section \ref{sec:Backbone}, \ref{sec:high_res_features} and \ref{sec:Feature_enhancement} respectively.

\subsection{Backbone Model} \label{sec:Backbone}
As shown in section (a) (highlighted red) of Fig.~\ref{fig:overview} we build upon a generic backbone action recognition model (see section \ref{sec:models} for specific model selection) to extract features from the frame level across the input clip and produce a final action classification. This backbone operates on spatially downsized video voxels of size $\mathbb{R}^{T,3,224,224}$ where T is the number of 224x224 RGB image frames, producing frame level feature maps, $\mathcal{F} \in \mathbb{R}^{T,C,H,W}$, which we seek to enhance via high resolution hand features.

\subsection{High Resolution Hand Feature Extraction} \label{sec:high_res_features}

Shown in section (b) (highlighted green) of Fig.~\ref{fig:overview} the second section of our model is designed to extract features from around the hands in order to aid the classification of actions with fine-grained spatial detail and is a two stage process. 

First, to detect the presence and location of an assemblers hand within an image we train a custom mobilenetv2 \cite{mobnetv2} model to predict the probability and location of hands within an image. As our model relies on the accurate localisation of an assemblers hands within video frames we report the high performance of our hand model in section A of the supplementary material along with model architecture, implementation and training details.

Second, a crop of size $(h_c,w_c)$ is taken centred on hand locations retaining the resolution of the full input image\footnote{If no hand is predicted by the model we input a central crop from the full image, which we found to perform better than a zeroed input.}. We truncate the hand models to extract features from an intermediate layer, producing two hand feature maps, $\mathcal{F}_h \in \mathbb{R}^{T,C’,H’,W’}$ in order to retain spatial features and limit parameters used within these network streams (see section \ref{sec:depth_choice} where we discuss our specific model and layer choices).

\subsection{Feature Enhancement} \label{sec:Feature_enhancement}

At the core to our model is the feature enhancement stage whereby extracted hand features maps ($\mathcal{F}_h$) are used to enhance the backbone model features ($\mathcal{F}$). We introduce the feature enhancement strategy highlighted by (c) (highlighted yellow) in Fig.~\ref{fig:overview} providing the models and dimensions of feature maps we select in section \ref{sec:depth_choice}. Our feature enhancement strategy allows for the spatial alignment and combination of features of any size from the hand stream to the features at any point in the backbone architecture.

\subsubsection{Spatial Alignment}

Due to the higher relative resolution of the hand input, compared to the frame wide input image, the hand input image and all hand feature maps ($\mathcal{F}_h$) extracted by the hand model correspond to a fixed region within the larger input frame and all subsequent feature maps $\mathcal{F}$ of the backbone model. In order to account for the large discrepancy in spatial input space between the hand and backbone stream, and to combine features in a spatially consistent manner we propose a spatial alignment process whereby features extracted from the high resolution hand data are mapped to their corresponding size and locations with respect to the input image via the use of up/down sampling zero padding. These padded feature maps represent the high resolution hand features with respect to the backbone architecture, which can be used to enhance the spatial features of the backbone feature maps. In order to achieve this spatial alignment between hand cutouts of size $(w_c$,$h_c)$ at location $(x_{hand},y_{hand})$, and the backbone randomly cropped input of size $C_s$ at an offset $(x_c$,$y_c)$, we calculate the normalised size $(\overline{w}_{hand},\overline{h}_{hand})$, and normalised offsets $(\overline{x}_{hand},\overline{y}_{hand})$ with in equation \ref{eqn:size} and \ref{eqn:offsets} respectively. Where H is the height of full resolution frame and S is the height of the scaled input image before random cropping -- a more detailed description and diagram are provided in section 2 of the supplementary material. 

\begin{equation}\label{eqn:size}
    \overline{w}_{hand} = \biggl(\frac{w_c}{H}\biggr)\left(\frac{S}{C_s}\right), \quad  \overline{h}_{hand} = \biggl(\frac{h_c}{H}\biggr)\left(\frac{S}{C_s}\right)   
\end{equation}

\begin{equation}\label{eqn:offsets}
    \overline{x}_{hand} = \frac{x_{hand}-x_c}{C_s}, \quad  \overline{y}_{hand} = \frac{y_{hand}-y_c}{C_s}
\end{equation}

Once the normalised offsets and size have been calculated, $\mathcal{F}_h$ is up/downsampled (using nearest neighbour when upsampling) to spatially match the size of the hand with respect to back bone feature map $\mathcal{F}$, to produce an intermediate feature map $\mathcal{F}_h’ \in \mathbb{R}^{T,C,H*\overline{h},W*\overline{w}}$. We then zero pad around $\mathcal{F}_h'$ to match the size of $\mathcal{F}$ using the offsets $(\overline{x}_{hand},\overline{y}_{hand})$ to place the features at the corresponding location within $\mathcal{F}$ producing a set of feature maps $\mathcal{F}’ \in \mathbb{R}^{T,C,H,W}$ matching the backbone spatial dimensions whilst containing high resolution hand features spatially located at the position of hands within the backbone feature map $\mathcal{F}$.

\subsubsection{Feature Combination}

Once hand features have been spatially aligned using zero padding to match the spatial dimension of the full frame features, they are concatenated along the channel dimension, producing a set of separate but spatially matched backbone and high resolution hand features $\in \mathbb{R}^{T,3C,H,W}$.

To combine these separate features into one feature set of enhanced full frame feature maps a 1x1 convolution, of kernel size C is utilised to reduce the channel dimension from 3C back to that of the original extracted feature map $\mathcal{F}$. This convolution allows for spatially preserved channel mixing between the feature maps and produce a set of enhanced hand feature maps. The enhanced feature maps are then added to the original extracted frame features, $\mathcal{F}$, via a residual connection, followed by a linear activation function and batch normalisation.

\subsection{Model \& Feature Combination Stage Selection} \label{sec:depth_choice}

Whilst our outlined method has so far has been invariant to model choice hand crop size and the intermediate feature maps selection, we now outline the choices made when implementing our model for the experiments described in section \ref{sec:results}.

\subsubsection{Model Selection} \label{sec:models}

Following convention we opt for a ResNet-50\cite{resnet} backbone and ResNet-18 models for the two hand streams. However note other more efficient backbones \eg \cite{mobnet} \cite{efficientnet} could be utilised for real time operation on edge devices. We also select a hand crop size of $w_c = h_c = 224$ and the random spatial crop $C_s = 224$ as this is the standard input size for a ResNet model, and simplifies equation \ref{eqn:size} to $\overline{w} = \overline{h} = \frac{S}{W}$.

We operate models across multiple frames utilising the Temporal Segment Network\cite{TSN} approach of averaging frame levels predictions over an input clip. To facilitate the efficient learning of temporal features between frames, we utilise the Temporal Shift Module\cite{TSM} which is inserted in every residual block for all models.

\subsubsection{Combination Stage}

When selecting the feature maps to utilise for feature combination, we note that minimal downsampling/upsampling is preferred when matching the two feature sets. As such we utilise the consecutive downsampling of the hand streams to the point where the downsampled hand features match the size of the hand within one of the early feature map within the backbone model (\ie  $H\overline{h} \approx H’$). We therefore opt to enhance the features from the first block of a ResNet-50 backbone $(\mathcal{F} \in \mathbb{R}^{T,256,56,56})$ with the features from a truncated hand models at the penultimate residual block $(\mathcal{F}_h \in \mathbb{R}^{T,256,14,14})$, as the backbone hands have a size of $H\frac{S}{H} = 56*(\frac{256}{720}) \approx 20$ and $H' = 14$.

\subsection{Real Time Action Segmentation} \label{sec:window}

We extend our offline action recognition approach to action segmentation via a simple overlapping sliding window methodology. During inference we utilise a dense frame sampling strategy to artificially create input segments corresponding to the current video frame $x_0$. T frames are sampled at a temporal stride of $\tau$ going backwards from the most recently received video frame to create an input clip of fixed length $\{x_{-T\tau},x_{-(T-1)\tau},...,x_{-\tau},x_0\}$ which is used to create a classification for the middle frame of the clip, $\ie$ $x_{-T\tau/2}$.

\subsubsection{Training Clip Creation - Surround Sampling}

To match the dense sampling strategy utilised at inference, during training clips of length $T\tau$ must be sampled from examples within our dataset containing N frames. To ensure our model can accurately determine the start and end of actions in sequences within a video feed we implement a surround sampling procedure where the starting frame of the training sample ($\ie$ $x_{-T\tau}$) is chosen via equation \ref{eqn:sample}, where $N_s$ and $N_e$ are the segment level labelled start and end frame of the segment within the dataset, T is the number of frames selected and $\tau$ is the temporal stride between frames.

\begin{equation} \label{eqn:sample}
    x_{-T\tau} = U\{N_s-\frac{T\tau}{2} , N_e-\frac{T\tau}{2}\}, s.t.\ \ x_{-T\tau} \in \mathbb{N}
\end{equation}

Importantly, this sampling procedure allows frames from outside the labelled segment to be seen by the model during training. As the model produces a prediction for the central frame of the segment this sampling strategy assists the model at accurately predicting the boundaries of actions much better than sampling strategy where training clips are created from data fixed within the bounds of an action.

During offline testing clips are sampled from the centre of the N frame examples, with the initial frame selected as $x_{-T\tau} = \frac{N_f - N_s}{2} - \frac{T\tau}{2}$.

\subsubsection{Temporally Aware Label Cleaning} \label{sec:clean}
Under the outlined framework, our model is used to produce per frame classification when operating in an overlapping sliding window fashion on sequences of video data. We implement an additional temporally aware label cleaning stage to improve the model performance on sequences of data as a method of reducing over segmentation while retaining real time implementation rather than using an offline method such as encoder/decoder. While the model is running across a sequence of data any predictions which are statistically too short, that is have a length shorter than $C_{mean} - \kappa C_{std}$, where $C_{mean}$ and $C_{std}$ are the mean and standard deviation of the lengths of class C within the training data, and $\kappa$  is a constant which we find by sweeping between 1 $\rightarrow$ 2 on the training data to determine the best cutoff for performance, are ignored by the model with the previous action prediction being taken as true.

\section{Experimental Results} \label{sec:results}

The following section presents the experimental results of our enhanced hand feature model on our newly collected atomic assembly dataset, comparing to baseline and similar works. We further discuss the deployment of our model (and others) for real time action segmentation through the use of temporally aware label cleaning, comparing the model to more computationally heavy encoder/decoder methods, highlighting the importance of short term temporal feature learning for fine-grained atomic action segmentation.

\subsection{Atomic Assembly Dataset} \label{sec:dataset}

\begin{table}[tb]
\label{tab:data}
\begin{center}
\centering
\scalebox{0.63}{

\begin{tabular}{m{0.1cm}m{1.5cm}m{3cm}m{2.9cm}m{1cm}m{0.75cm}m{0.75cm}}
\toprule
\textbf{i.d} & \textbf{Interaction} & \textbf{Object} & \textbf{Secondary Object} & \textbf{N$^o$ of Examples} & \textbf{Mean Length (sec)} & \textbf{Total Length (min)} \\ \midrule
0 & Unbox & Component & - & 82 & 8.63 & 11.80 \\ \hline
1 & \multirow{5}{*}{Pick Up} & Washer and/or Screw & - & 610 & 2.07 & 21.08 \\
2 &  & Spanner & - & 50 & 1.44 & 1.20 \\
3 &  & Screwdriver & - & 101 & 1.02 & 1.73 \\
4 &  & Marker Pen & - & 149 & 1.12 & 2.78 \\
5 &  & Torque Wrench & - & 85 & 1.14 & 1.62 \\ \hline
6 & \multirow{4}{*}{Put Down} & Spanner & - & 49 & 0.96 & 0.78 \\
7 &  & Screwdriver & - & 102 & 1.13 & 1.92 \\
8 &  & Marker Pen & - & 140 & 1.27 & 2.96 \\
9 &  & Torque Wrench & - & 81 & 1.07 & 1.49 \\ \hline
10 & \multirow{5}{*}{Place} & Washer and Screw & on Screwdriver & 428 & 1.86 & 13.29 \\
11 &  & Washer and Screw & on Product & 159 & 1.78 & 4.72 \\
12 &  & Component & on Product & 123 & 3.43 & 7.03 \\
13 &  & Washer & on Bolt & 74 & 2.76 & 3.40 \\
14 &  & Metal Bar & on Product & 38 & 3.59 & 2.27 \\ \hline
15 & \multirow{6}{*}{Tighten} & Screw & with Hand & 163 & 3.80 & 10.32 \\
16 &  & Screw & with Screwdriver & 359 & 5.32 & 31.85 \\
17 &  & Screw & with Torque Wrench & 168 & 3.39 & 9.49 \\
18 &  & Nut & with Hand & 77 & 8.43 & 10.82 \\
19 &  & Nut & with Spanner & 220 & 1.99 & 7.28 \\
20 &  & Nut & with Torque Wrench & 95 & 7.74 & 12.26 \\ \hline
21 & \multirow{2}{*}{Mark} & Bolt & with Marker Pen & 134 & 4.46 & 9.97 \\
22 &  & Screw & with Marker Pen & 105 & 12.89 & 22.55 \\ \hline
23 & Remove & Washer and Nut & from Product & 76 & 7.95 & 10.07 \\ \hline
24 & \textbf{No Action} & - & - &  \textbf{1220} & \textbf{3.75} & \textbf{76.26} \\ \bottomrule
\end{tabular}
}
\caption{\small Summary of assembly dataset statistics.} \label{tab:data}
\end{center}
\end{table}

In order to evaluate the effectiveness of our model for HRC tasks we create a novel manufacturing action segmentation dataset. A D415 Intel RealSense depth camera (15 fps @ 920x720) was used over 3 days to record 3 different workers complete a real world robotics assembly procedure. The camera was positioned above a workbench with the camera angled to capture the bench surface, workers hands and lower torso, avoiding discernible features. Assemblers were informed to perform assemblies as normal, making no alterations for the camera. In total 38 full assemblies were recorded, corresponding to $\sim$6 hours of footage.

\textbf{Labelling.} The data is labelled using a fine-grained noun/verb labelling procedure with class selection guided by the standard operating procedure for the given assembly. We select fundamental interactions between the worker and either the product being assembled, a part/component or a tool - while ignoring unuseful or ambiguous actions for understanding the assembly process. In total seven primary interactions (verbs) were found: \textit{Unbox, Pick Up, Put Down, Place, Tighten, Mark with Pen, Remove}. These interactions either occurred with a single object or with a primary and secondary object. In total 24 relevant actions verb/noun pairs were found between the operator and all objects within the assembly and are summarised in Table \ref{tab:data}. To label the data we utilise a custom action segmentation labelling GUI, which we make public\footnote{https://github.com/Matthewkm/Action-Segmentation-Labeller-GUI}. In total 186 632 from a total of $\sim$324 000 frames ($\sim$57.6$\%$) were found to contain an atomic action from our label list.

\textbf{Properties.} Our dataset follows a long tail distribution with the number of collected examples ranging from 38 (\textit{Place Metal Bar} - only occurring once per assembly) to 610 (\textit{Pick up Washer and/or Screw}) with average clip lengths ranging from as short as 0.96 seconds (\textit{Put Down Spanner}) to 12.89 seconds (\textit{Mark Screw(s) with Pen}). Total clip lengths of classes range from as low as 46 seconds (\textit{Put Down Spanner}) to over 30 minutes (\textit{Tightening Screw with Screwdriver}). While relatively modest in size our dataset is emblematic of many dataset collected within real world environments where the collection and labelling of large datasets is challenging due to quick turnaround times and domain novelty. Compared to other action segmentation datasets ours is much more fine-grained than large datasets such as Breakfast~\cite{Breakfast} (77 hours - 10 actions) and larger in size than comparative fine-grained datasets such as GTEA~\cite{gtea} (28 minutes - 20 actions) and 50 salads~\cite{50Salads} (4 hours - 17 actions). Our dataset is noticeably fine-grained as the background remains constant throughout all classes, with only subtle changes in spatial and temporal information between many classes. In addition, due to the noun/verb pairing process there are overlaps between some classes where the same noun (e.g tool or component) appears within multiple verb pairings (e.g screw or place) and vice versa. As such the ability to model both spatial and temporal information will be key to accurate classification.

\begin{table}[]
\begin{center}
\centering
\scalebox{0.8}{
\begin{tabular}{l|c|ccc}
\toprule
\multicolumn{1}{c|}{\multirow{2}{*}{Model}} & Segment & \multicolumn{3}{c}{Sequence Segmentation} \\
\multicolumn{1}{c|}{} & Level F$_1$ & \multicolumn{3}{c}{F$_1$@\{0.5,0.25,0.1\}}\\ \midrule
Single Frame Baseline & 68.6 & 30.7 & 44.9 & 51.7 \\
TSM \text{\cite{TSM}} ($\ie$ backbone only) & 86.2 & 67.3 & 79.8 & 81.3 \\
TSM$_{HA}$ \cite{hand_attention} & 79.7 & 57.7 & 70.2 & 72.3 \\
TSM$_{EHF}$ (ours) & \textbf{89.2} & \textbf{69.2} & \textbf{80.5} & \textbf{82.5} \\ \midrule
 Reduced Resolution  TSM$_{EHF}$ & 86.1 & 67.2 & 78.2 & 79.9   \\
High Res Hands Only & 87.6 & 67.1 & 78.6 & 80.4 \\
\bottomrule
\end{tabular}
}
\end{center}
\caption{\small Comparison of model performance on both a segment level and on unseen extended assembly sequences utilising temporally aware label cleaning.}\label{tab:results}
\end{table}

\subsection{Implementation Details}
All models are trained within the Pytorch deep learning framework on segments extracted from 25 of the full assembly sequences, reserving the remaining 13 assembly sequences for unseen testing on both a segment and sequence level. See the supplementary information for a detailed account of parameter settings when training the models.

\subsection{Evaluation Metrics}
In order to test our models spatial and temporal reasoning we report results on both pre-cropped segments as well as temporal segmentation on unseen sequences. We report $F_1$ score on a segment level treating each pre-cropped segment as an single example regardless of length. When analysing segmentation results on an unseen sequence we report frame wise accuracy (Acc) as well as segmental edit distance, and segmental F1 score (as proposed by \cite{TCN}) at various IoU thresholds (10\%, 25\% and 50\%) to accurately account for over-segmentation errors.

\subsection{Evaluation of High Resolution Hand Model} \label{sec:hand_results}

\begin{figure}[t]
    \centering
    \includegraphics[width=0.5\textwidth]{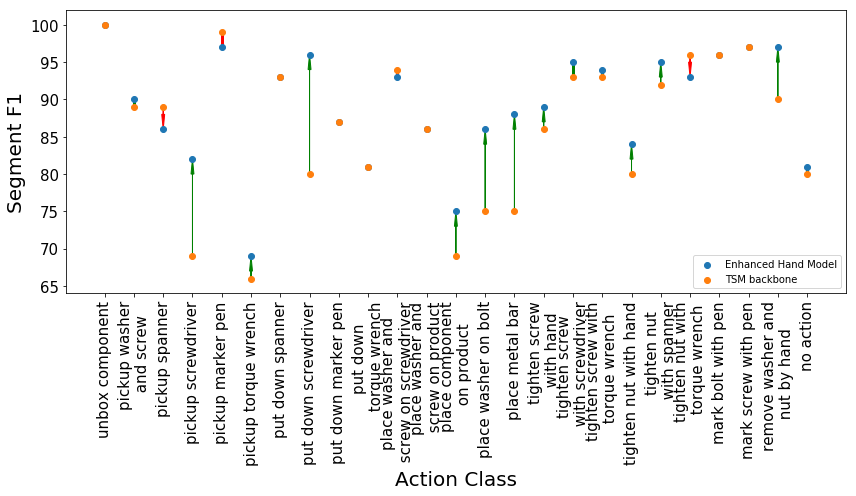}
    \caption{\small Improvement in F1 score across classes on a segment level when comparing an 8 frame temporally aware TSM model to an equivalent model with our high resolution hand enhanced features.}
    \label{fig:hand_seg}
\end{figure}

Table \ref{tab:results} compares the performance of our enhanced hand feature model (TSM$_{EHF}$) to an equivalent unenhanced TSM \cite{TSM} model (both T=8 frames, with a temporal stride, $\tau$=8) on a segment and extended sequence level. We further compare our model to an implementation of the similar hand attention (TSM$_{HA}$) work by Kobayashi $\etal$\footnote{We note that Kobayashi $\etal$ operate in an encoder/decoder structure, however, we re-implement their work following our method outlined in section \ref{sec:window} as well as utilise a temporally aware backbone structure to match our work.} \cite{hand_attention} and a baseline single frame ResNet-50 model. We find our model outperforms the backbone TSM model and the hand attention model by Kobayashi.

Fig. \ref{fig:hand_seg} highlights how TSM$_{EHF}$ improves the per class performance for nearly all classes compared to an equivalent backbone model, with the largest improvement seen in many classes with fine grained spatial differences, such as distinguishing between \textit{place washer on bolt} and \textit{place metal bar}. Our model does not harm the temporal reasoning as we see no decrease in temporally salient classes, and even see large improvements in \textit{pick up/put down screwdriver} and \textit{tighten nut with hand} vs \textit{remove washer and nut by hand}, which require both fine-grained spatial and temporal reasoning. Only a small improvement is seen for \textit{No Action}, suggesting our model is not improving the ability to distinguish between \textit{No Action} and an action (as we see going from temporally unaware to temporally aware models - see section \ref{sec:temporal}) but is extracting complementary enhanced spatial features, helping to distinguish between fine-grained spatial classes, without harming a models ability to temporally model frame wide features.

\begin{figure}
    \centering
    \includegraphics[width=0.5\textwidth]{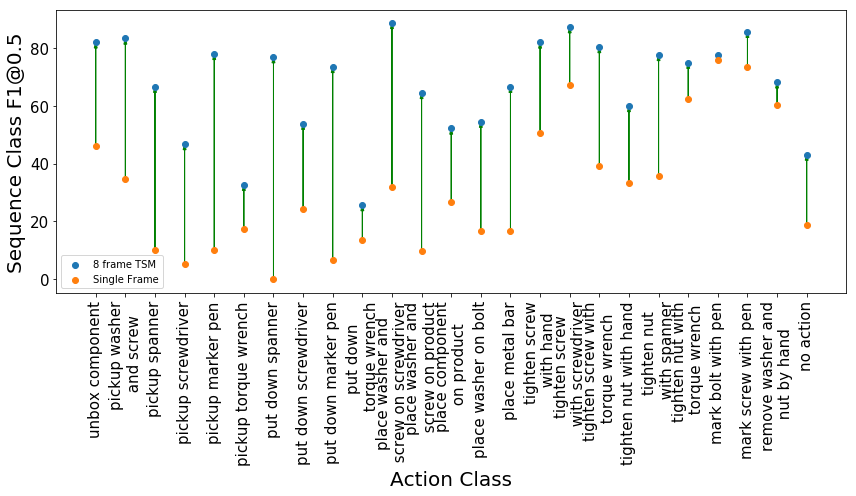}
    \caption{\small Improvement in $F_1@0.5$ score across classes on a sequence level when comparing a single frame to an 8 frame temporally aware TSM model operating in a sliding window fashion.}
    \label{fig:seq_classes}
\end{figure}

To further verify the performance improvements of TSM$_{EHF}$ are due to high resolution hand data otherwise not available to the baseline TSM model, we implement two modified models also shown in Table \ref{tab:results}. Firstly, just the high resolution hand section of our model (implementing the full ResNet-18 models with averaged softmax outputs) finding that the while the model performs better than the backbone TSM model on a temporally cropped segment level (despite having many more parameters) it performs worse on a sequence level, suggesting the larger frame wide features of the backbone model are useful for accurate segmentation under our framework ($\ie$ distinguishing boundaries of actions). Secondly, a low resolution hand version of our model, where hand input images are downsampled to match the resolution of the input image to the backbone model, essentially creating a model with the same number of parameters, but without increased resolution. As TSM$_{EHF}$ outperforms both these models and the TSM baseline on a segment level we can see that high resolution hand features provide separate yet complementary features useful for classification, and that it is the access to higher resolution features that improves classification performance.

\subsection{Real Time Action Segmentation Results}\label{sec:AS}

\begin{table*}[]
\centering
\scalebox{0.85}{
\begin{tabular}{l|c|c|c|l|lll|l}
\toprule
Segmentation Approach & Model & Finetuned From & Real time & Acc & \multicolumn{3}{l}{F1@\{0.5,0.25,0.1\}} & Edit \\ \midrule
\multirow{3}{*}{\begin{tabular}[c]{@{}l@{}} Simple Sliding Window \end{tabular}} & TSN \cite{TSN} & Kinetics & \cmark & 74.1 & 24.9 & 34.6 & 39.5 & 31.2\\
& TSM \cite{TSM} & Kinetics &\cmark & 81.2 & 47.9 & 56.1 & 57.8 & 44.5\\
& TSM$_{EHF}$ (ours) & Kinetics & \cmark & \textbf{81.8} & 49.1 & 56.5 & 58.9 & 45.4 \\ \midrule
\multirow{3}{*}{\begin{tabular}[c]{@{}l@{}} Temporally Aware \\ Label Cleaning (ours) \end{tabular}} & TSN \cite{TSN} & Kinetics & \cmark & 73.6 & 49.1 & 67.8 & 72.9 & 65.4 \\
& TSM \cite{TSM} & Kinetics &\cmark & 79.7 & 74.0 & 84.6 & 85.6 & 77.7 \\
& TSM$_{EHF}$ (ours) & Kinetics & \cmark & 80.9 & \textbf{75.8} & \textbf{85.6} & \textbf{87.2} & \textbf{79.9} \\ \midrule
\multirow{3}{*}{Encoder/Decoder \cite{ms-TCN}} & ms-TCN & TSM Kinetics & \xmark & 70.7 & 51.2 & 68.5 & 73.8 & 71.4 \\
& ms-TCN & TSM Assembly Dataset&\xmark & \textbf{84.1} & 77.6 & 85.3 & 86.3 & 79.6 \\
& ms-TCN & TSM$_{EHF}$ Assembly Dataset & \xmark & 83.7 & \textbf{78.4} & \textbf{87.0} & \textbf{88.4} & \textbf{83.6} \\ \bottomrule
\end{tabular}
}
\caption{\small Sequence Segmentation results for temporally aware label cleaning of sliding window predictions and ms-TCN using background emitted performance metrics as utilised in \cite{ms-TCN}. All models operating on T=8 frames with temporal stride $\tau$=8 } \label{tab:segment_results}
\centering
\end{table*}

In the following section we compare the performance of temporally aware label cleaning as a method of deploying AR models to unseen extended sequences of assembly data. We show fine-grained temporally aware backbones are key to successful classification of short atomic actions and that temporally aware label cleaning is capable of matching the performance of encoder/decoder methods on temporally short atomic actions while maintaining the ability to operate in a near real time manner.

\vspace{-4mm}
\subsubsection{Importance of Short Range Temporal Learning} \label{sec:temporal}

One key design choice within our model framework is that of the backbone model for classification and specifically how important is the ability for the backbone model to learn temporal features across multiple frames. Under our dense sampling strategy outlined in \ref{sec:window} there are two variables facilitating how much temporal information the backbone model can access from an input clip: the number of input frames, T, and the stride between them, $\tau$. In order to investigate the importance of temporal reasoning we train a set of identical Temporal Segment Networks (TSN) with and without inserted TSM with varying numbers of input frames and temporal stride between input frames, with detailed results shown in section D of the supplementary material. Through our evaluation we find, for atomically short actions, that a temporally aware backbone operating on multiple frames is key to accurate classification on a sequence level. We find that performance saturates at T=8,$\tau$=8 (real time FPS of 1.875 with the model only having access to 4.27 seconds of video) when temporally aware label cleaning is utilised.

Fig. \ref{fig:seq_classes} shows the improvement in individual class performance when changing from a single to an 8 frame temporally aware backbone model when performing action segmentationn in sliding window fashion. When utilising a single frame model many classes achieve very poor performances especially the more temporally salient classes, such as the \textit{pick up/put down tool} and \textit{place} classes, as a single frame model can't distinguish between picking up a tool and holding a tool (which falls under no action). As no action is a more prominent class, in the absence of the ability to distinguish between these actions and no action the model ends up classifying all these actions as no action. Once temporal reasoning has been enabled all classes are substantially improved, with some such as \textit{put down spanner} seeing an improvement from 0 to 78 $F_1@0.5$. We also observe that the improvement in classification of temporally salient classes and no action is also much larger on a sequence level compared to a segment level (see supplementary information) suggesting a temporally aware backbone is key for accurate action segmentation when operating in a sliding window fashion, allowing the model to distinguish between actions and no action much more successfully.

\vspace{-0.5cm}
\subsubsection{Importance of Temporally Aware Label Cleaning}\label{sec:cleaning_results}

The first two sections of Table \ref{tab:segment_results} show the performance before and after temporally aware label cleaning (section \ref{sec:clean}) of various models operating in a sliding window fashion on unseen assembly sequences from our fine-grained assembly dataset. Again we see that enhanced hand features continue to improve the model performance. Whilst we only see a modest change in frame level accuracy we observe that temporally aware label cleaning significantly improves the performance of our models on a sequence level, while allowing the models to maintain their near real time implementation\footnote{There is a small delays in model inference due to the need for T$\tau$/2 future frames for prediction and $<$T$\tau$/2 frames for cleaning - however this delay is $<$2 second for a 15 fps camera)}. This intuitively suggests temporally aware label cleaning is combating over-segmentation errors, which is further supported by the fact that as the short term temporal modelling ability of a model increase ($\ie$ TSN $\rightarrow$ TSM) the effects of the temporally aware label cleaning also increase, however much less of a change is observed for the enhanced spatial feature models ($\ie$ TSM $\rightarrow$ TSM$_{EHF}$).

\vspace{-3mm}
\subsubsection{Comparison to Encoder/Decoder Methods}

Finally we compare our temporally aware label cleaning method to the commonly used encoder/decoder multi-sage Temporal Convolution Network (ms-TCN) \cite{ms-TCN}. Table \ref{tab:segment_results} shows temporally aware label cleaning out performs a kinetics pretrained ms-TCN (utilising the same feature extraction model), while being able to operate in a near real time manner. We posit this is due to the fact that actions within our dataset are typically much shorter in nature than many other action segmentation datasets and do not necessarily occur in a fixed order, thus removing the advantage of extended temporal reasoning provided by an encoder/decoder method. We further show that features extracted from our assembly dataset trained models can be used to as base features to train ms-TCN models, providing significant improvements over the kinetics feature extraction (F$_1$@0.5 51.2 $\rightarrow$ 78.4) while reducing training time by $\sim$65\% (~200$\rightarrow$70 epochs) showing that significant improvements can be obtained for other encoder/decoder methods via the use of in domain pretraining via the methods outlined in section \ref{sec:window}.

\section{Conclusion}
In this paper we present a novel approach to utilising high resolution image features around an assemblers hand to improve action segmentation within real world assembly video. We show that through the use of short range temporally aware backbones, surround sampling and temporally aware label cleaning our model can be applied to extended unseen video sequences in a sliding window fashion, proving real time action segmentation required for many assembly HRC tasks. Further work will focus on applying models to other similar domain publicly available datasets as well as a more generalised network capable of high resolution features extraction from self determined areas of interest for use outside hand orientated datasets.

{\small
\bibliographystyle{ieee_fullname}
\bibliography{egbib}
}

\newpage\hbox{}\thispagestyle{empty}\newpage
\appendix

\section{Hand Localisation}

\subsection{Model details}
In order to detect the presence and location of an assemblers hand within an image we train a mobilentv2 \cite{mobnetv2} model to output a fixed Sigmoid normalised vector of length 6 containing the elements $[\hat{P}_1,\hat{x}_1,\hat{y}_1,\hat{P}_2,\hat{x}_2,\hat{y}_2]$ where $\hat{P}_1$ and $\hat{P}_2$ are the probability of the left and right hand existing in the image (1 if hand present 0 otherwise) and $\hat{x}_i$, $\hat{y}_i$ are the normalised position of hand i in a given input image. We train with a modified mean squared error loss, shown in equation \ref{eq:loss}, that doesn't penalise the model for incorrectly predicting a hand position when no hand is present, where $P_i$ is the (1,0) label of a present hand and $x_i$ and $y_i$ are the ground truth normalised location of hand i within a given image. $\lambda$ is an equalising constant set to 0.1 to get similar learning rates across the loss function. 

\begin{align*}\label{eq:loss}
\mathcal{L}  &= \lambda \sum_{i=1}^{2}(P_i-\hat{P}_i)^2 + P_1[(x_{1}-\hat{x}_{1})^2 + (y_{1}-\hat{y}_{1})^2 ] \\
    &\hskip5em\relax + P_2[(x_{2}-\hat{x}_{2})^2 + (y_{2}-\hat{y}_{2})^2 ] \numberthis 
\end{align*}

\subsection{Implementation Details}
Our hand localisation algorithms were trained with a learning rate of 0.04 for 200 epochs with a learning rate decrease at 100 and 150 epochs. With gradient clipping set to 20, parameter weight decay set to 0.0005 and momentum set to 0.9.

We train our hand model on 4531 randomly selected frames within our training data and test on a further 1919 frames. Roughly 15\% of selected frames contain either just one or no hands. We labelled our hand data with a custom GUI which we also plan to make publicly available.

\subsection{Hand Localisation Performance}

In order to evaluate the performance of of hand localisation model we implement the following metric: $F1@T_L$$>$$X$. Under our model output we get a prediction per hand in the form $(\hat{P},\hat{x},\hat{y})$ where $\hat{P}$ is the probability of a hand being present within the frame, and $\hat{x},\hat{y}$ are the normalised position of the hand within the frame. We allocate our predictions a true positive label if $\hat{P}$ is grater than 0.5 ($\ie$ there is more than 50\% probability the hand is present) and the there is a ground truth hand present, and the spatial prediction of the hand location is within a certain threshold of the ground truth location, given by $\sqrt{(\hat{x}-x)^2 + (\hat{y}-y)^2} < T_L$, where $T_L$ is a changeable threshold and $x$ and $y$ are the ground truth hand position. If a hand is predicted to exist but has the wrong location or there is no hand present then the prediction will be assigned false positive. A prediction is false negative when a hand not predicted but a hand is present within the image. We report the hand performance for various location thresholds $T_L$ in table \ref{tab:hand_performance}.

\begin{table}[]
\begin{center}
\begin{tabular}{l|c}
\toprule
$T_L$ & $F1@T_L$  \\ \hline
0.05 & 74.5        \\
0.1  & 83.0        \\
0.2  & 92.5        \\
0.3  & 96.8        \\ \bottomrule
\end{tabular}
\end{center}
\caption{Hand localisation performance on our novel assembly dataset.} \label{tab:hand_performance}
\end{table}

\section{Hand Feature Spatial Alignment}

\begin{figure}
    \centering
    \includegraphics[width=0.5\textwidth]{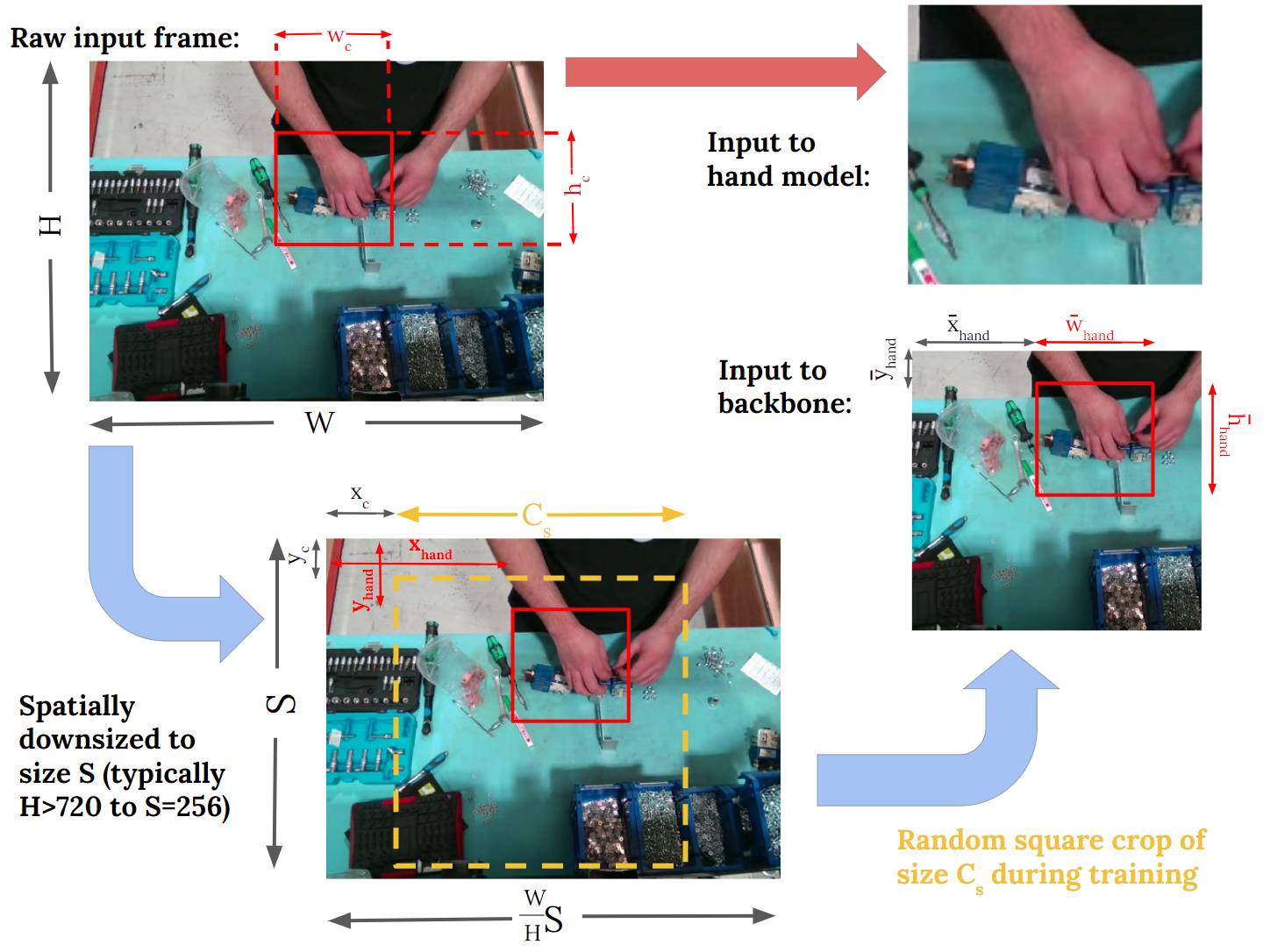}
    \caption{The location and size of the extracted hand image with respect to the input image to the backbone model after pre-processing.}
    \label{fig:hand_dims}
\end{figure}

Assuming an input image of size (W,H) and a down sample operation to size S for input into the back bone model, the relative resolution between the two inputs is $(H/S)$. In order to achieve this spatial alignment between hand cutouts of size $(w_c$,$h_c)$, and the backbone cutout of size $C_s$, we calculate the normalised size $(\overline{w}_{hand},\overline{h}_{hand})$ via equation \ref{eqn:size} and offsets $(\overline{x}_{hand},\overline{y}_{hand})$ via equation \ref{eqn:offsets} with respect to the input image to the backbone stream. The variables in these equations are defined in Fig. \ref{fig:hand_dims} which shows the down scaling of the image from the shortest side from $H$ to $S$ and random crop of size $C_s$ at offset $(x_c,y_c)$ (for regularisation during training) of the input frame. 

\section{Model Implementation Details}
All models were implemented within the Pytorch deep learning framework with a Resnet50 model used for the backbone architecture and a ResNet18 used for the hand model. All models utilised Kinetics-400\cite{kinetics} pretraining, with batch normalisation statistics frozen from pretrained weights to reduce overfitting. The initial learning rate was set to 0.002 for parameters within the backbone model and 0.0002 for parameters within the hand feature extraction model, with gradient clipping set to 20. Parameter weight decay was set to 0.0005 and momentum set to 0.9 for all parameters. All models were trained for 150 epochs with a batch size of 64 (except for the 16 frame model which utilised a batch size of 32), with learning rate decreased by a factor of 10 at 100 and 125 epochs. A dropout of 0.8 was implemented in the final fully connected layer of all model to reduce overfitting. 

Models were trained on the a GPU node of the Hartree Centre Jade-2 HPC \footnote{https://www.jade.ac.uk/}, utilising a single Tesla V-100 GPU with a wall clock training time of $\sim$ 6 hours per model.

\section{Importance of Temporal learning}

\begin{figure}
\hspace*{-0.5cm} 
\begin{tabular}{cc}
  \includegraphics[width=0.5\linewidth]{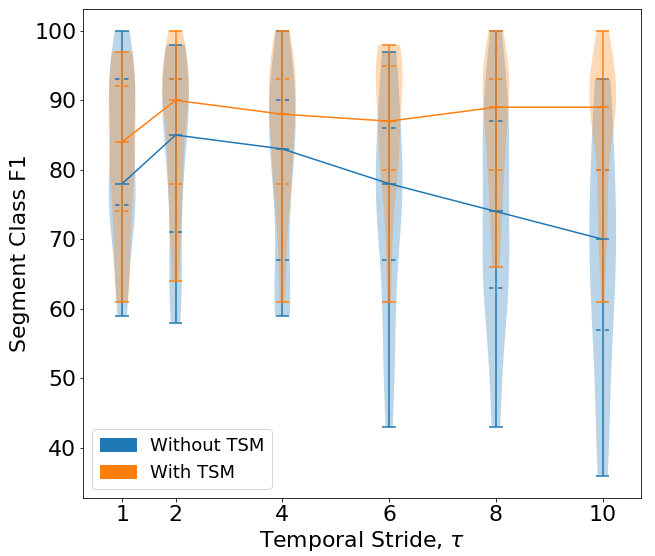} &   \includegraphics[width=0.5\linewidth]{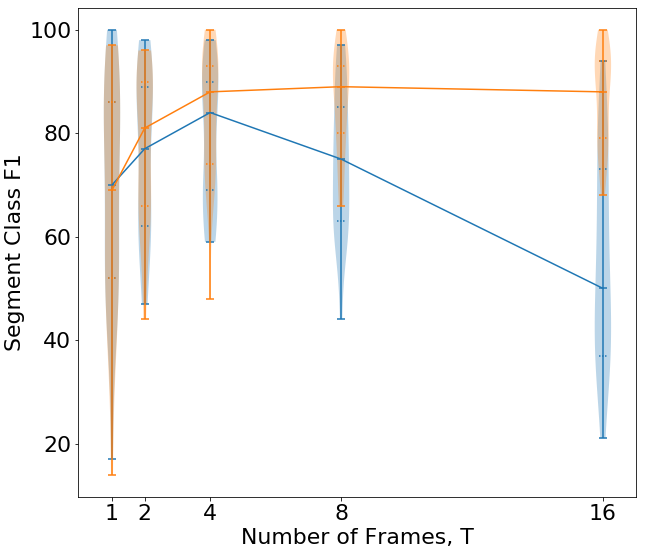} \\
 (a) & (b) \\[6pt]
 \includegraphics[width=0.5\linewidth]{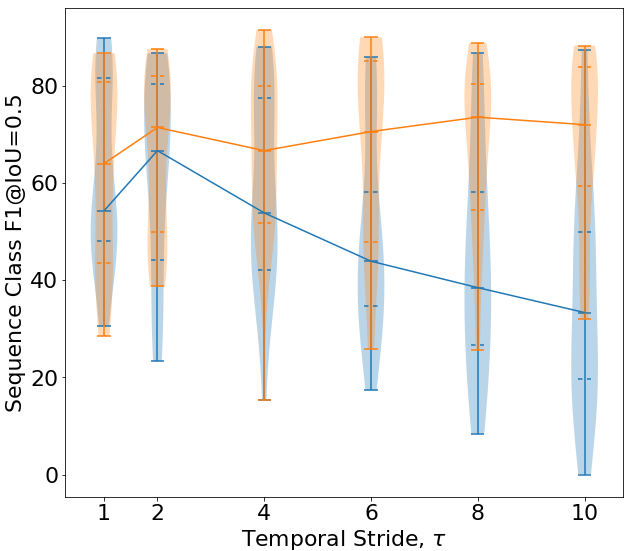} \label{fig:seq_stride} &   \includegraphics[width=0.5\linewidth]{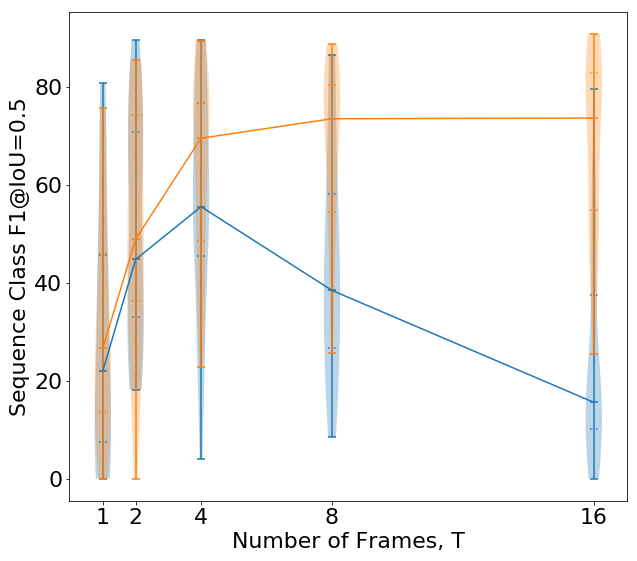} \label{fig:seq_frames}\\
(c) & (d) \\[6pt]
\end{tabular}
\caption{\small \textbf{Importance of temporal reasoning in the backbone model.} All graphs show the F1 class distributions for a temporal segment network \cite{TSN} with and without inserted temporal shift modules\cite{TSM} varying either the temporal stride, $\tau$, with fixed T=8 (a and c), or varying number of input frames, T, with fixed $\tau$=8 (b and d) on either a segment (a and b) or on an extended sequence (c and d) level after temporally aware label cleaning.}\label{fig:temporal}
\end{figure}

In order to investigate the importance of temporal reasoning we train a set of identical Temporal Segment Networks (TSN) with and without inserted TSM with varying numbers of input frames and temporal stride between input frames, with results shown in Fig. \ref{fig:temporal}.

We first fix T to the commonly used 8 frames \cite{IBM,TSM} and vary the temporal stride $\tau$$\in $\{1,2,4,6,8,10\} with results shown in Fig. \ref{fig:temporal} (a and c on a segment and sequence level respectively). It is clear that temporally aware TSM models outperform standard TSN models across all temporal strides, and maintain a high F1 score beyond $\tau=2$, while the performance of a regular TSN model, incapable of temporal learning, drops as $\tau$ increases, suggesting sparser frames can introduce more useful long term information for classification, but only when temporal learning is possible. This drop in performance is more significant on a sequence level with the performance of some classes dropping to an F1 score of 0 at $\tau=10$, suggesting sparser frames when temporal learning is not applicable leads to significant confusion when operating in a sliding window fashion.

Secondly, we vary the number of input frames  T$\in$\{1,2,4,8,16\} while keeping $\tau=8$. Fig.~\ref{fig:temporal} (b and d on a segment and sequence level respectively) show that for both a TSN and TSM model increasing the number of input frames helps the models initially, however, beyond $T=4$ the TSN model performance drops significantly, while the TSM model continues to perform well, suggesting again that providing more temporal information by using multiple frames at input is only useful if a temporally aware backbone model is used.

\section{Single \textit{vs.} 8 Frame - Segment Level}

\begin{figure}[h]
    \centering
    \includegraphics[width=0.5\textwidth]{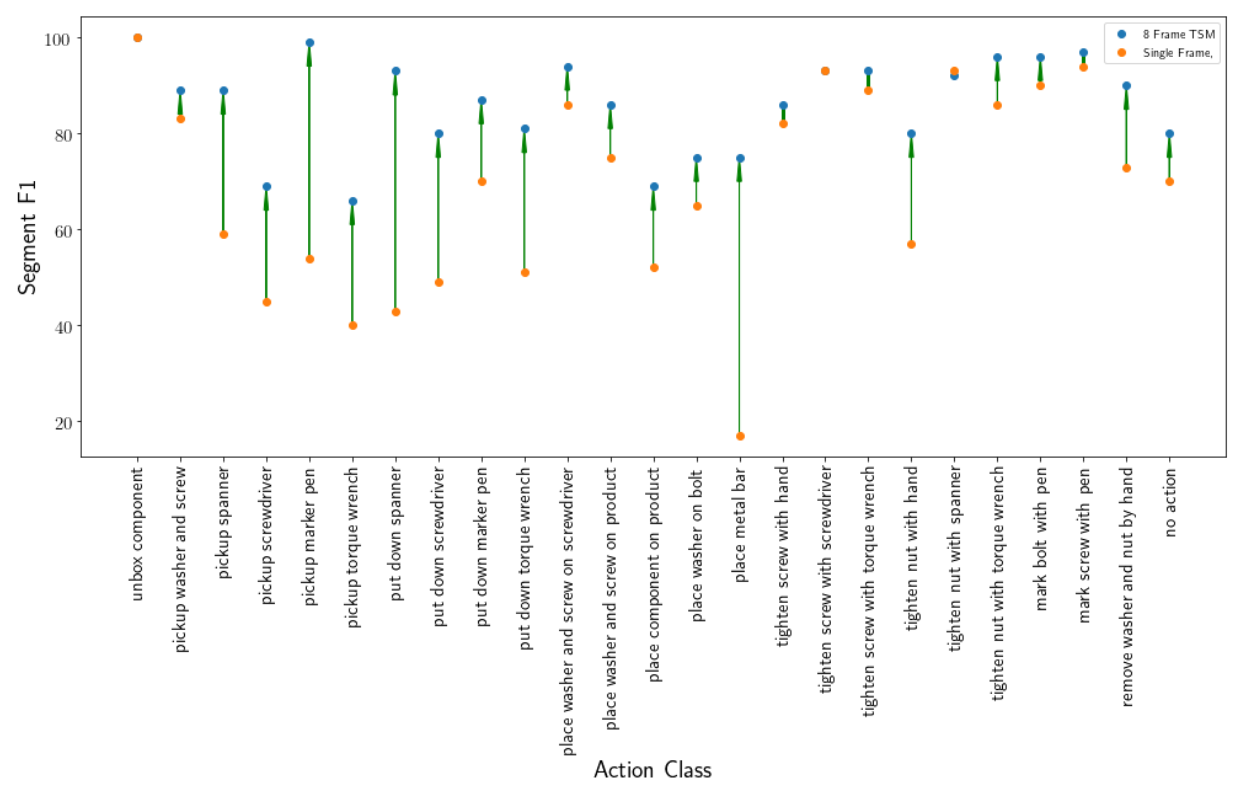}
    \caption{Improvement in $F1$ score across classes on a segment level when comparing a single frame to an 8 frame temporally aware TSM model.}
    \label{fig:seg_8frame}
\end{figure}

Figure \ref{fig:seg_8frame} shows the class F1 improvements comparing a single frame and a temporally aware 8 frame model on a temporally cropped segment level. In keeping with analysis on a sequence level, temporally salient classes see the largest improvement, with place metal bar seeing the largest improvement from an F1 score of 17 to 79. It is noted that the improvements in class performance on a segment level is smaller than that on a sequence level as mentioned in section 1.4.4, suggesting temporal modelling is more imperative when operating on a sequence level, with no action seeing a much smaller improvement, suggesting when segments are neatly temporally cropped they are easier to distinguish on a spatial level.

\section{High Resolution Hand model - Sequence Level}
\begin{figure}[ht]
    \centering
    \includegraphics[width=0.5\textwidth]{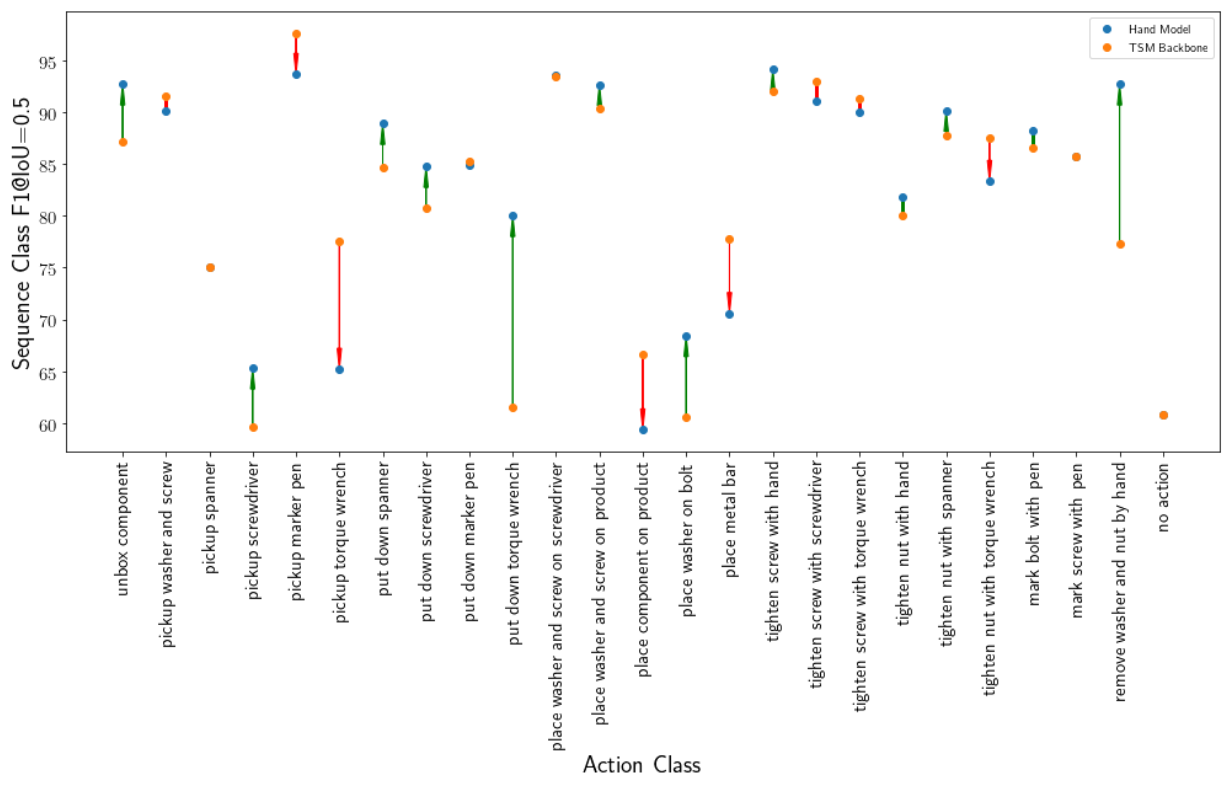}
    \caption{Improvement in $F1@IoU>0.5$ score across classes on a sequence level when comparing an 8 frame temporally aware TSM model to an equivalent model with our high resolution hand enhanced features.}
    \label{fig:seq_hand}
\end{figure}

For completeness we also include the difference in class performance between a baseline TSM model and our high resolution hand model on sequence level. Figure \ref{fig:seq_hand} shows that there is virtually no change in no action performance on a sequence level, again suggesting high resolution hand features are improving the models ability to distinguish between classes rather than distinguish between an action and no action.

\section{Model Prediction Visualisations}

\begin{sidewaysfigure*}
    \centering
    \includegraphics[width=\textwidth,height=\textheight,keepaspectratio]{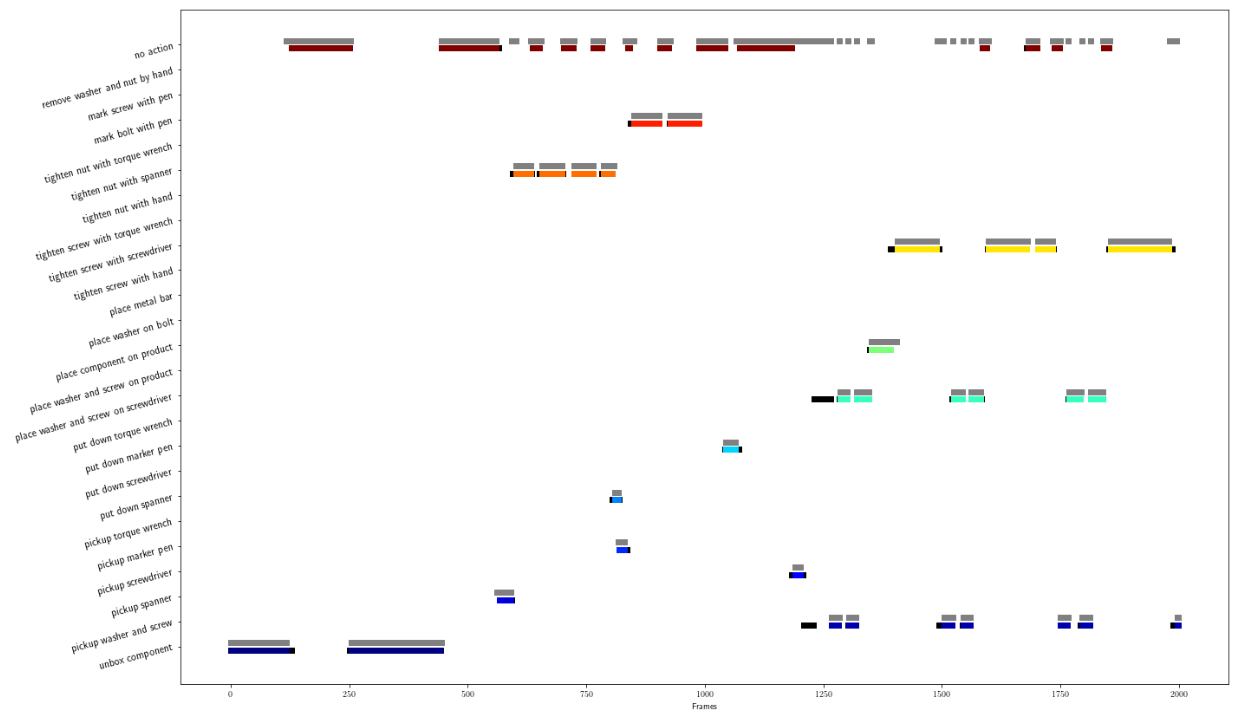}
    \caption{Model predictions on an extended video sequence.}
    \label{fig:seq1}
\end{sidewaysfigure*}

\begin{sidewaysfigure*}
    \centering
    \includegraphics[width=\textwidth,height=\textheight,keepaspectratio]{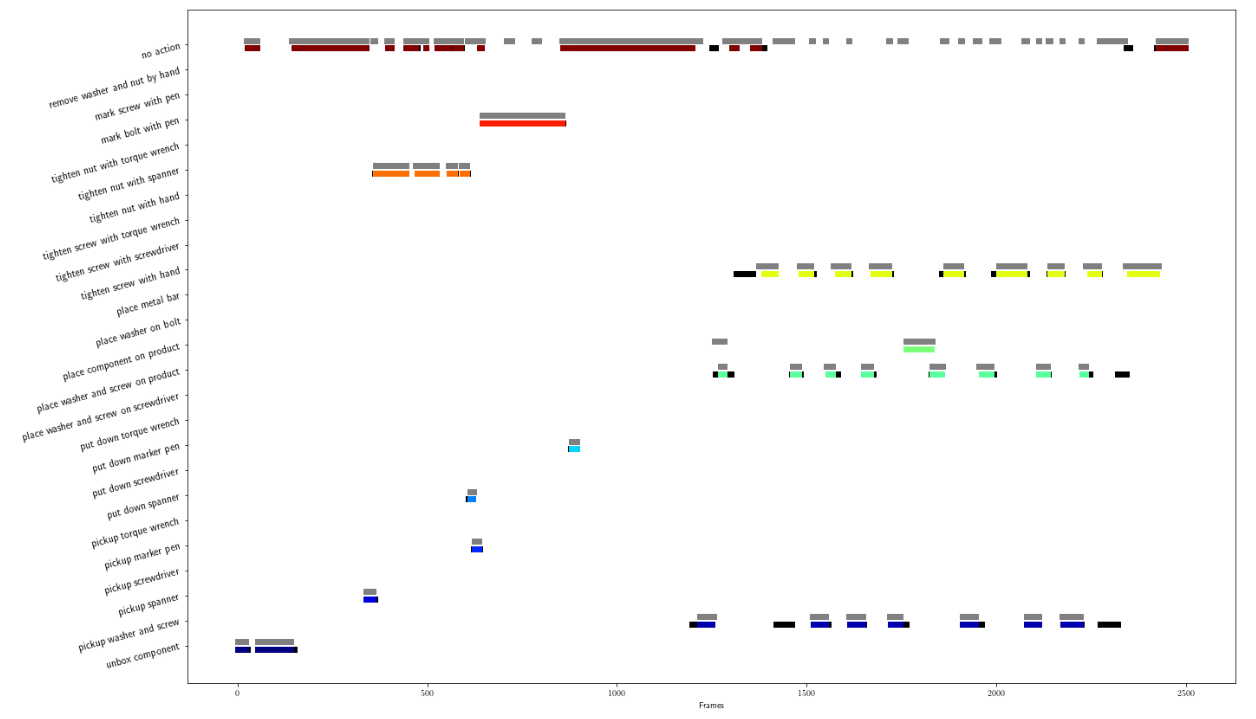}
    \caption{Model predictions on an extended video sequence.}
    \label{fig:seq2}
\end{sidewaysfigure*}

In the following section we produce two model prediction visualisations extracted from two extended unseen assembly sequences using an 8 frame TSM model with high resolution hand feature enhancement. Figure \ref{fig:seq1} and \ref{fig:seq2} show class colour coded predictions against ground truth predictions in grey, with black predictions representing incorrectly predicted frames.

As can be seen, despite the model operating in a sliding window fashion it is capable of accurately classify nearly all actions, predicting the start and end of actions successfully despite not being explicitly trained to do so. One notable challenging scenario encountered by the model is rapid changes from one action to another which are often separated by a few frames of ``no action". This problem is highlighted in Figure \ref{fig:seq2} when the operator repeatedly places a screw on the product and then tightens the screw with their hand. In this situation the model is liable to miss many short no action segments - combining them into one of the rapidly changing actions, however can still accurately segment the respective actions with very good precision.

\end{document}